\pdfoutput=1

\documentclass[11pt]{article}
\usepackage[dvipsnames]{xcolor}
\usepackage{acl}

\usepackage{times}
\usepackage{latexsym}

\usepackage[T1]{fontenc}

\usepackage[utf8]{inputenc}

\usepackage{microtype}

\usepackage{inconsolata}

\usepackage{graphicx}

\usepackage{microtype}
\usepackage{hyperref}
\usepackage{url}
\usepackage{multirow}
\usepackage{microtype}
\usepackage{hyperref}
\usepackage{url}
\usepackage{multirow}

\usepackage{xcolor,pifont}
\newcommand*\colourcheck[1]{%
  \expandafter\newcommand\csname #1check\endcsname{\textcolor{#1}{\ding{52}}}%
}
\colourcheck{ForestGreen}

\usepackage{caption}
\usepackage{subcaption}

\newcommand{\dataset}{\textsc{NESTful}}

\definecolor{light-gray}{gray}{0.9}

\usepackage{float}

\usepackage{bbm}
\usepackage{todonotes}
\usepackage{multirow, makecell}
\usepackage{arydshln}
\usepackage{booktabs}
\usepackage{amsmath}

\usepackage{listings}
\usepackage{xcolor}
\colorlet{punct}{red!60!black}
\definecolor{background}{HTML}{EEEEEE}
\definecolor{delim}{RGB}{20,105,176}
\colorlet{numb}{magenta!60!black}
\usepackage{multirow}
\usepackage{colortbl}
\usepackage[T1]{fontenc}
\usepackage[]{hyphenat}

\lstdefinelanguage{json}{
    basicstyle=\scriptsize\ttfamily,
    numbers=none,
    numberstyle=\scriptsize,
    stepnumber=1,
    numbersep=8pt,
    showstringspaces=false,
    breaklines=true,
    frame=lines,
    backgroundcolor=\color{background},
    literate=
     *{0}{{{\color{numb}0}}}{1}
      {1}{{{\color{numb}1}}}{1}
      {2}{{{\color{numb}2}}}{1}
      {3}{{{\color{numb}3}}}{1}
      {4}{{{\color{numb}4}}}{1}
      {5}{{{\color{numb}5}}}{1}
      {6}{{{\color{numb}6}}}{1}
      {7}{{{\color{numb}7}}}{1}
      {8}{{{\color{numb}8}}}{1}
      {9}{{{\color{numb}9}}}{1}
      {:}{{{\color{punct}{:}}}}{1}
      {,}{{{\color{punct}{,}}}}{1}
      {\{}{{{\color{delim}{\{}}}}{1}
      {\}}{{{\color{delim}{\}}}}}{1}
      {[}{{{\color{delim}{[}}}}{1}
      {]}{{{\color{delim}{]}}}}{1},
}

\title{\dataset{}: A Benchmark for Evaluating LLMs \\ on Nested Sequences of API Calls}

\author{
  Kinjal Basu$^{1*}$,
  Ibrahim Abdelaziz$^{1*}$,
  Kiran Kate$^1$, 
  Mayank Agarwal$^1$, 
  Maxwell Crouse$^1$, \\ 
  {\bf Yara Rizk$^1$, 
  Kelsey Bradford$^2$, 
  Asim Munawar$^1$, 
  Sadhana Kumaravel$^1$, 
  Saurabh Goyal$^1$,} \\ 
  {\bf Xin Wang$^1$,
   Luis A. Lastras$^1$, and 
  Pavan Kapanipathi$^1$}  \\
  $^1$IBM Research, USA  \hspace{10mm} $^2$Georgia Institute of Technology \\
  \normalsize{\{kinjal.basu, ibrahim.abdelaziz1, mayank.agarwal, maxwell.crouse, yara.rizk\}@ibm.com} \\
  \normalsize{\{asim, sadhana.kumaravel1, saurabh.goyal1, chloe.wang \}@ibm.com} \\
  \normalsize{kelseymaria5@gmail.com, \{kakate, lastrasl, kapanipa\}@us.ibm.com}
}

\begin{document}

\maketitle
\def\thefootnote{*}\footnotetext{These authors contributed equally to this work}\def\thefootnote{\arabic{footnote}}
\begin{abstract}

The resurgence of autonomous agents built using large language models (LLMs) to solve complex real-world tasks has brought increased focus on LLMs' fundamental ability of tool or function calling. 
At the core of these agents, an LLM must plan, execute, and respond using external tools, APIs, and custom functions. Research on tool calling has gathered momentum, but evaluation benchmarks and datasets representing the complexity of the tasks have lagged behind. 
In this work, we focus on one such complexity, nested sequencing, with the goal of extending existing benchmarks and evaluation. Specifically, we present \dataset{}
, a benchmark to evaluate LLMs on nested sequences of API calls, i.e., sequences where the output of one API call is passed as input to a subsequent call. \dataset{} contains 1800+ nested sequences where all the function calls are executable.
Experimental results on a variety of models show that the best-performing model (GPT-4o) achieves a full sequence match accuracy of 28\% and a win-rate of 60\%, necessitating a large scope for improvement in the nested sequencing aspect of function calling.
Our analysis of these results provides possible future research directions for the community, in addition to a benchmark to track progress. 
We have released the \dataset{} dataset under the Apache 2.0 license at \url{https://github.com/IBM/NESTFUL}.
\end{abstract}

\section{Introduction}


Autonomous agents, built with Large language models (LLMs), are gaining popularity in solving complex, real-world problems \cite{yaoreact,deng2024mind2web}. LLMs handle a user's request by understanding their intents, planning the required tasks to address it, executing those tasks step by step, and providing a response. For most real-world problems~\cite{jimenezswe,roy2024exploring,thakur23language}, LLMs must interact with external environments through tool, function, and API calls (Application Programming Interface), which primarily leverages LLMs' tool calling abilities\footnote{API, function calling and tool-use are used interchangeably throughout the paper}.

The significant reliance on LLMs' function calling abilities led recent research to continuously improve this dimension of LLMs. On one hand, approaches to improve function calling have exploded~\cite{abdelaziz2024granite,liu2024apigen,srinivasan2023nexusraven}; on the other hand, benchmarks and evaluations are lagging behind. For instance, BFCL v1 and v2 focused on evaluating single, multiple, and parallel function calling tasks for both non-executable and executable versions~\cite{berkeley-function-calling-leaderboard}. Works such as API-Blend~\cite{basu2024apiblendcomprehensivecorporatraining} complemented prior work by introducing granular task evaluation such as slot-filling, API-detection, and sequencing. BFCL v3 has progressed further into agentic use cases with multi-step and multi-turn function calling evaluation. 

\begin{figure*}[t!]
\begin{center}
\includegraphics[width=\textwidth]{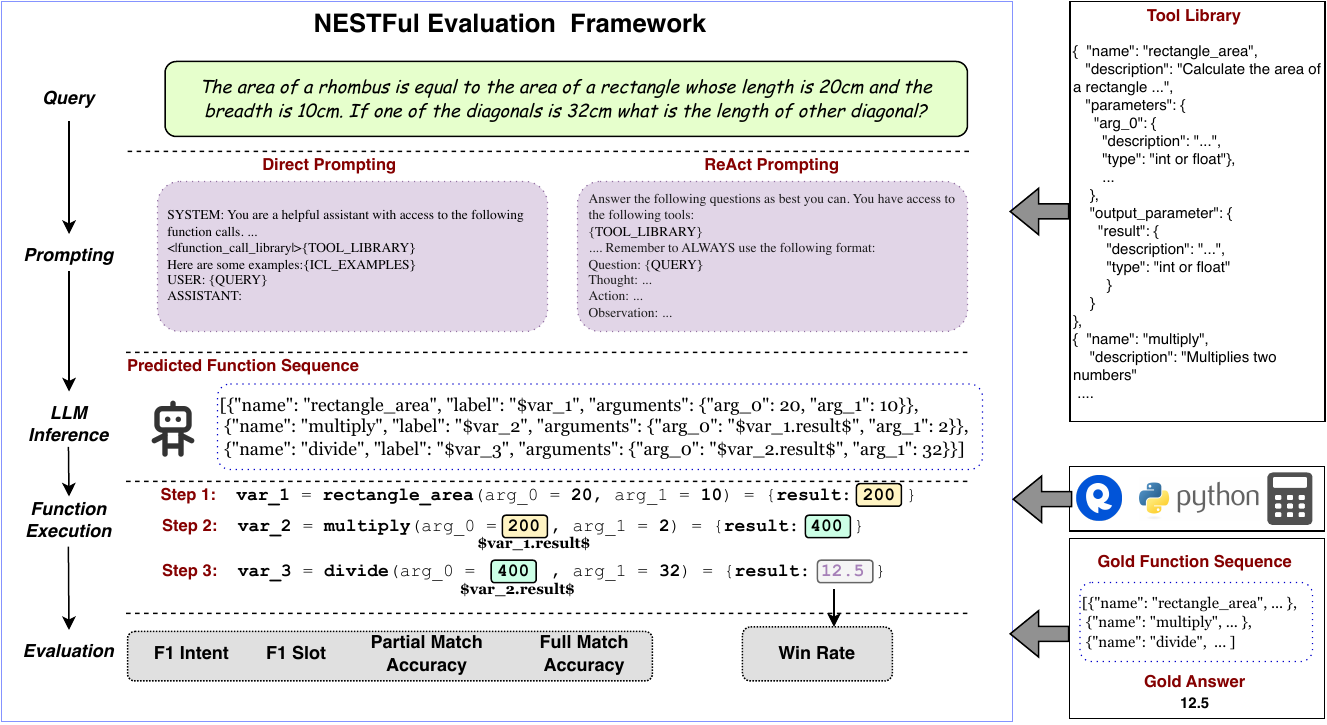}
\caption{End-to-End Evaluation Pipeline: \dataset{} provides a test set of input queries and its corresponding list of nested function calls. It also provides executable implementations for each tool in the library and allows for evaluating models in direct prompting or REACT styles. Given a query, the pipeline infers the input LLM to generate the required sequence of function calls, execute those functions (taking into account nested variables) and compare the final answer with the gold one. }
\label{fig:data_flow}
\end{center}
\end{figure*}

However, benchmarks for fundamental but complex tasks such as the sequencing of functions have not been well explored yet, which forms the basis of this work. Existing evaluation benchmarks pose sequencing as the prediction of single or multiple isolated API calls, where the output of any particular API call within that sequence is considered irrelevant. In contrast, for many real-world tasks, a sequence of API calls is nested, i.e., the output of some API calls is used in the arguments of subsequent API calls. Figure \ref{fig:data_flow} shows one such example of a nested sequence of APIs.

In this paper, we present \dataset{}, a benchmark specifically designed to evaluate models on nested API calls with over 1800 nested sequences. It consists of: (1) user queries, (2) a catalog of APIs and their specifications, (3) the sequence of API calls and the corresponding parameters, and (4) the expected output response. The datasets are based on the MathQA~\cite{mathqa} and StarCoder2-Instruct~\cite{wei2024selfcodealign} datasets which are commonly used in the literature but are missing the executable component.  Figure \ref{fig:data_flow} shows the end-to-end evaluation pipeline of \dataset{}.

We evaluated the dataset on 19 state-of-the-art models from the literature and exposed the gaps of these models in handling complex function calling sequences. GPT-4o acheived the best performance, but did not exceed 28\%  full sequence match nor 60\% on win rate metrics. Models struggled as the nesting got deeper and the data dependencies increased. To further advance research in this area, we will publicly release the \dataset{} dataset with executable Python implementation for each tool and the evaluation code for all models.

\section{Related Work}


The best way to enable API function calling in LLMs remains an active area of research. Methods that utilize large, general-purpose proprietary models (e.g., Gemini \citep{team2023gemini} or GPT \citep{achiam2023gpt}) typically make use of carefully constructed prompts and in-context learning examples, e.g., \cite{song2023restgpt}. Smaller, more specialized models often start from a strong-performing code model (e.g., DeepSeek-Coder \citep{guo2024deepseek}, CodeLlama \citep{roziere2023code}, or Granite Code \citep{mishra2024granite}) and fine-tune primarily on highly curated datasets \cite{srinivasan2023nexusraven, gorilla-openfunctions-v2, abdelaziz2024granite} that have been extended with synthetic data \cite{zhang2024agentohana}. 

Most of these existing works initially focused on basic function calling abilities that did not involve much complexity. Recent advances in enabling LLMs to handle complex multi-API interactions have introduced structured methods like Reverse Chain \cite{zhang2024reversechaingenericrulellms}. This approach employs backward reasoning to optimize multi-step API planning, allowing LLMs to effectively manage nested workflows by aligning intermediate steps with the final goal. Such methods highlight LLMs' potential to perform efficient, target-driven planning in resource-constrained environments.

To evaluate and enhance these aforementioned approaches, numerous works released training and benchmarking data in service of API function calling, such as ToolLLM \cite{qin2023toolllm}; APIBench \cite{patil2023gorilla}; APIGen \cite{liu2024apigen}; or API-BLEND \citep{basu2024apiblendcomprehensivecorporatraining}. While these benchmarks focus on simpler or isolated API calls, NesTools \citep{han2024nestoolsdatasetevaluatingnested} and our work target more complex, real-world tasks involving interdependent, nested tool use. Unlike the fully synthetic NesTools, \dataset{} is built from established datasets and has longer average call sequences (4.36 vs. 3.04). Other benchmarks like SealTool \cite{wu2024seal} and BFCL-v3 \cite{berkeley-function-calling-leaderboard} include some nesting but are smaller and not specifically designed for it.

ToolBench \cite{xu2023tool} includes 205 nested samples from the Webshop and TableTop datasets, compared to over 1800 in NestFul. Moreover, ToolBench supports only 34 APIs, whereas NestFul features more than 900 unique functions. ShortcutsBench \cite{shen2024shortcutsbench} has multi-step tool calls but is tailored specifically to Apple Shortcuts feature, with evaluation data generated in a proprietary format that is difficult to interpret and does not follow a standard JSON structure. Finally, AgentBoard is a broad evaluation framework for general-purpose agents for multi-step planning tasks with fine-grained metrics and visual tools, while \dataset{} focuses specifically on tool-augmented LLMs with detailed offline metrics like F1 score and accuracy alongside win rate.

\section{\dataset{} Dataset Curation}
\label{sec:dataset}

\dataset{} comprises of more than 1,800 instances designed for benchmarking tool calling in LLMs on nested sequencing. Each instance consists of (1) a user query, (2) a list of all available tools for the model to choose from, (3) the gold sequence of tools and their arguments needed to answer the user query, and (4) the final answer that should be obtained once the tools are executed. The dataset also contains corresponding Python code for every API in the library and a mechanism to run the input query via any LLM, execute the tools predicted by the LLM, and provide the final answer.

\begin{figure}[t]
    \centering
    \includegraphics[width=1.0\linewidth]{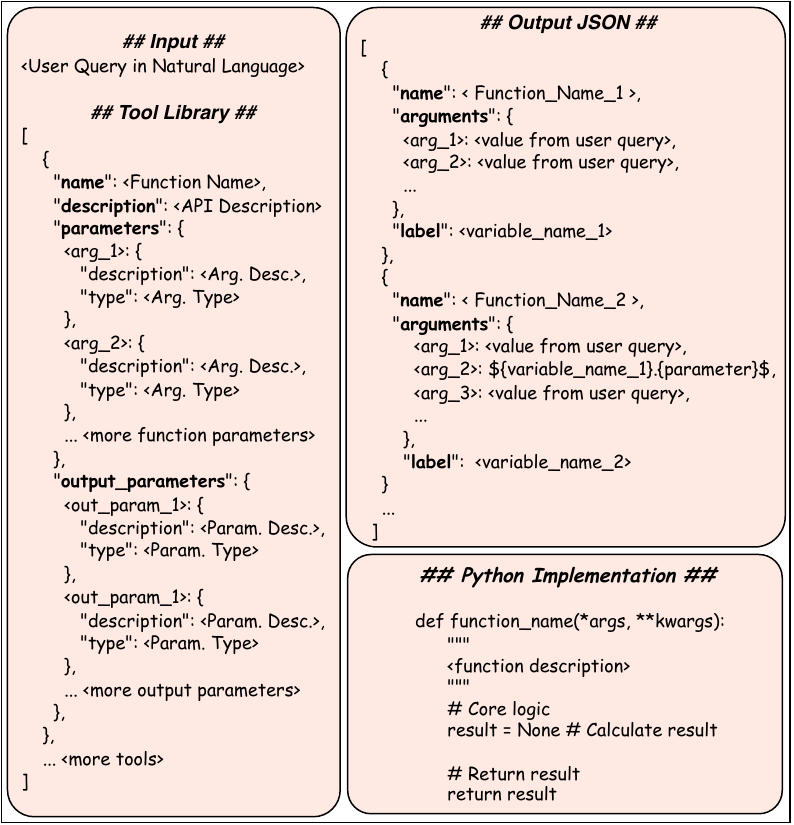}
    \caption{\dataset{} Data Schema for \textit{Input}, \textit{Tool Library}, \textit{Output}, and \textit{Python implementation}. In \textit{Output JSON}, the \texttt{arg\_2} of \texttt{Function\_Name\_2} showcases how the variable assignments are used to create a nested sequence of function calls.}
    \label{fig:data_schema}
\end{figure}

\subsection{Nested Function Calling Data Schema} \label{sec:data_schema}
\dataset{}'s data schema, demonstrated in Figure \ref{fig:data_schema}, showcases the template used for representing the \textit{Input}, \textit{Tool Library}, \textit{Output}, and \textit{Python Code}. An important aspect of nested function calling is to enable a mechanism for tool reference; i.e. a subsequent tool call using that reference to access the output of the previous tool execution. To do so, we assign a \textit{unique variable name} to each tool which distinctly identifies each tool, even when two identical tools with different arguments appear in the same sequence (parallel API calls). For example, in Figure ~\ref{fig:data_flow}, ``rectangle\_area'' tool was assigned ``label'': ``\$var\_1''. This allows the next tool ``multiply'' to use the output of ``rectangle\_area'' as an argument:  ``arg\_0'': ``\$var\_1.result\$''.

\subsection{\dataset{} Data Domains}
\dataset{} is composed of data from two domains; 1) mathematical reasoning data and 2) generic tools from the coding domain. We describe below the process followed to create each data category.

\begin{figure}
    \centering
    \includegraphics[width=\columnwidth]{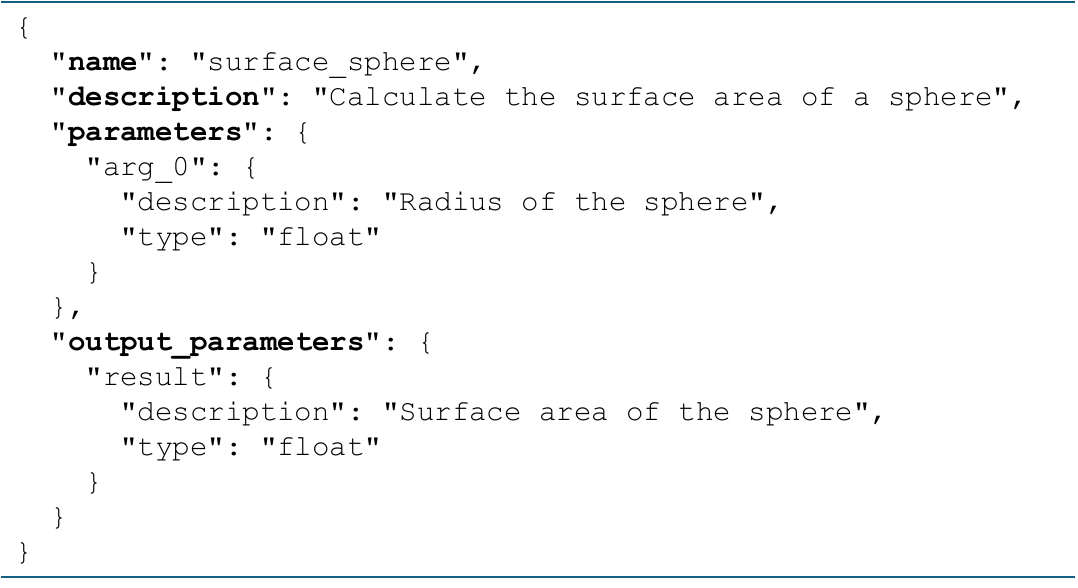}
    \caption{Sample specification for tools  from MathQA}
    \label{fig:mathqa_toolspec}
\end{figure}

\begin{figure*}
    \centering
    \includegraphics[width=0.8\linewidth]{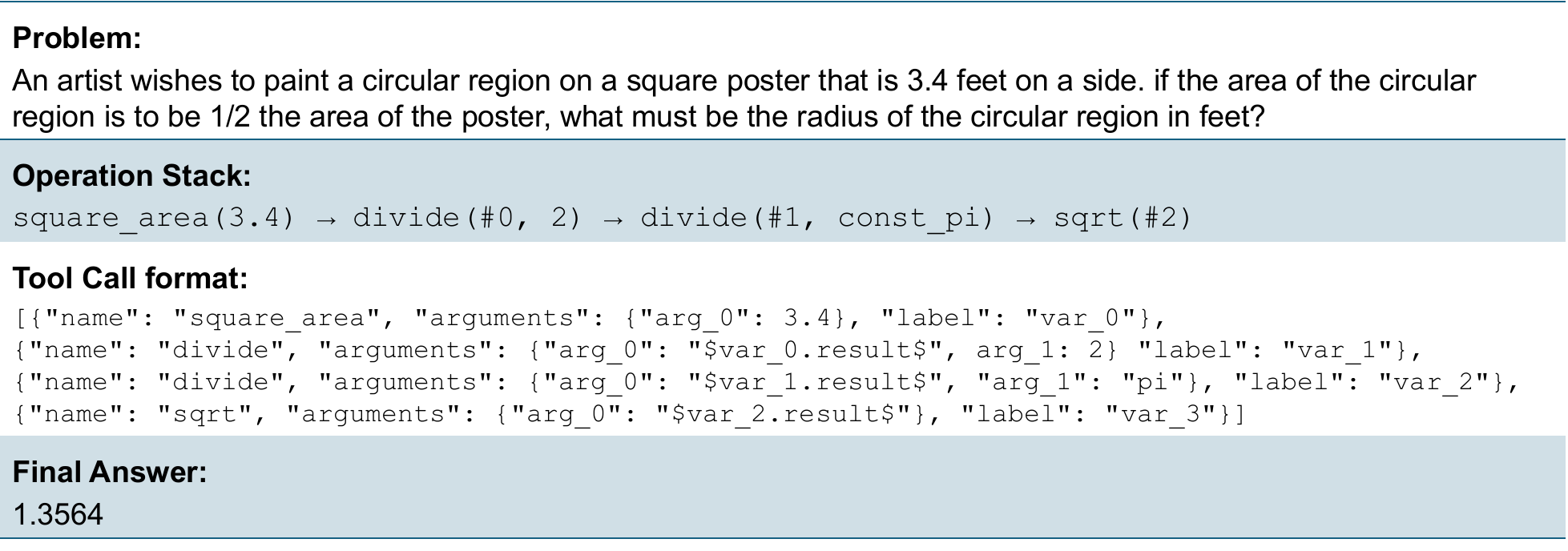}
    \caption{Sample problem from the MathQA dataset. The \textit{Operation Stack} provides an ordered sequence of nested tool calls which we transform into a \textit{Tool Call format} for the \dataset{} dataset.}
    \label{fig:mathqa_sample}
\end{figure*}

\subsubsection{Mathematical Reasoning Data}
For the first part of \dataset{}, we relied on datasets that test the model for nested function calling in the math domain. We build on MathQA~\cite{mathqa}; a benchmark designed to evaluate AI models' ability to solve mathematical word problems. It consists of questions that test numerical reasoning and problem-solving skills, requiring models to both understand the text of a word problem and perform mathematical operations to arrive at the correct solution. 

%
%
%
%

\paragraph{Tool specifications} Since MathQA provides only the tool names, we manually created specifications for all the tools in the dataset. This covers 40 tools in total; e.g.  divide, remainder, volume\_cylinder, permutation, etc. For each tool, we define the name, tool description, and detailed outline of the tool input and output parameters including the parameter data type and description as shown in Figure~\ref{fig:mathqa_toolspec}.


\paragraph{Tool calling input-output pairs} To build the test data, we used the test set of MathQA where the ``problem'' definition is the query and parsed the ``annotated formula'' into a nested sequence of tool calls. An example is shown in Figure \ref{fig:mathqa_sample}.




\paragraph{Executable Code and Filtering} For each tool, we also generated its corresponding implementation in Python. This allows us to execute the nested call sequence and match the execution result with the gold answer. It  also ensures the correctness of the set of corresponding tools and the code execution too. We then filtered out any samples where we could not reproduce the gold answer from executing the nested tool calls.
This process resulted in 1,415 test samples spanning 40 tools with an average 5.1 tool calls per sample.

\subsubsection{Coding Data}
We also curated test examples based on generic Python functions from the StarCoder2-Instruct dataset \cite{wei2024selfcodealign}. This dataset has a total of 50K Python functions and covers a wide range of tools that can be used. 
We started by collecting tool instructions and their Python implementations, followed by using Mixtral-8x22B to infer parameter type hints. Any functions that were syntactically incorrect or non-executable were filtered out. Next, a synthetic data generation pipeline was used to create instruction–nested call pairs using the valid seed tools and examples. This pipeline included a validator to ensure all parameters and tool names were accurate and complete. Finally, execution-based filtering was applied to verify that the generated samples produce the correct final output. We elaborate on each step in the following sections.

\begin{figure}[t]
    \centering
    \includegraphics[width=\columnwidth]{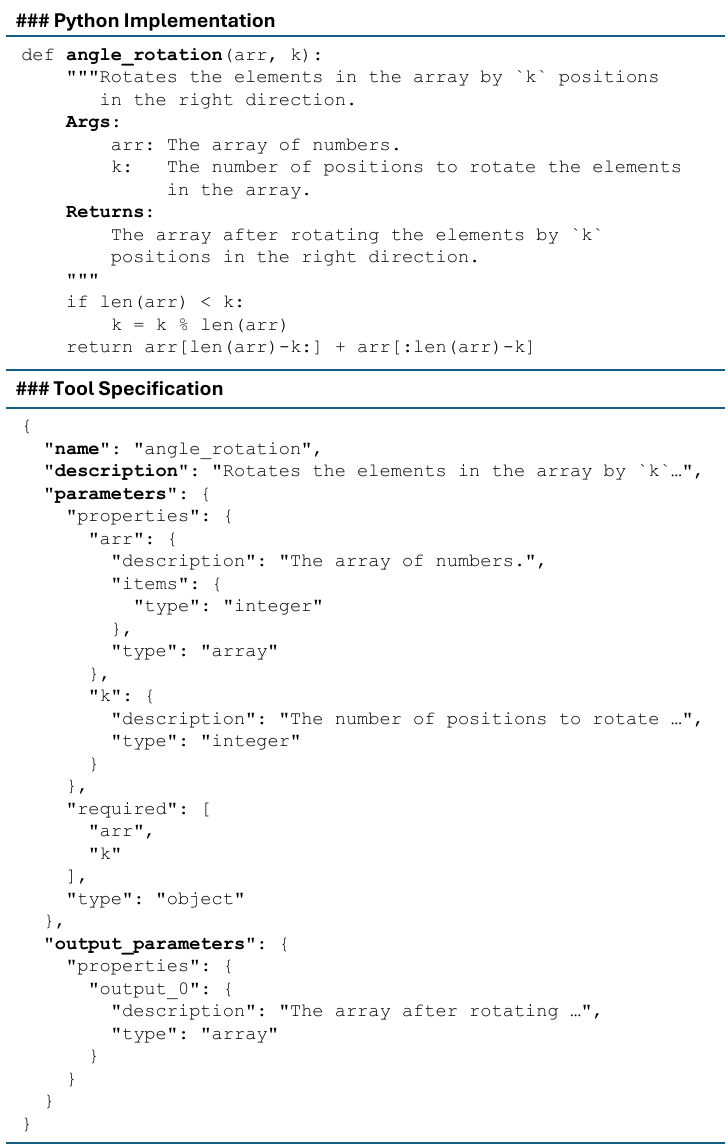}
    \caption{[\textit{top}] An example function (\textit{angle\_rotation}) from  StarCoder2-Instruct dataset with its docstring documentation. [\textit{bottom}] Tool specification for ``angle\_rotation'' after inferring the different data types and creating its input/output parameters.}
    \label{fig:stackqa_converted}
\end{figure}

\paragraph{Tool specifications} We leveraged StarCoder2-Instruct's Python implementations and docstrings to create API specifications, see Figure \ref{fig:stackqa_converted} for an example.  
For each Python function, (1) we used Mixtral 8x22B\footnote{\url{https://huggingface.co/mistralai/Mixtral-8x22B-Instruct-v0.1}} to generate the possible Python types of the input and output arguments, and (2) we validated and executed the function code itself to ensure it does not produce any errors. After both steps, we generated a corresponding JSON specification for each function documenting its input-output arguments. Figure \ref{fig:stackqa_converted} shows an example.

\paragraph{Tool calling input-output pairs} To create input-output pairs, we leveraged DiGiT\footnote{\url{https://github.com/foundation-model-stack/fms-dgt}} synthetic data generation framework. DiGiT allows for defining various synthetic data generation pipelines given seed examples of input/output pairs. In particular, we used 10 seed examples and used Mixtral-8x22B as the teacher model. We also implemented a function calling validator that applies various heuristics to check the quality of the synthetic data, ensuring function calls adhere to the given specifications. In particular, we have validations to ensure the tools and parameters used are not hallucinated, required parameters are specified, and there is at least one nested tool call in the output sequence.




\paragraph{Filtering} We further filtered the generated input-output pairs by executing their gold nested API sequence to ensure they execute and attach the result as the gold answer. For a function that has a randomness element (e.g., generating a random list), we set a fixed seed for all our experiments and re-execute all those cases to ensure that we are getting the same response all the time.
From this category of data, we generated 446 test examples covering more than 881 distinct tools with an average tool sequence length of 2.1.

\subsection{Dataset Quality}
Our benchmark builds on MathQA and StarCoder2-Instruct. MathQA is a well known mathematical reasoning dataset that was  manually validated by humans,  providing the input, correct sequence of math operations, and final answer. After converting it into a nested tool sequence, we further validate it by executing the sequence to ensure it produces the original gold answer. However, since the coding dataset is synthetically generated, we also implemented multiple automatic validation at various stages. This includes checking that nested tool sequences align with tool specifications and execute correctly to produce the expected output.

\section{Evaluation}
\label{sec:eval}

\begin{table*}[t]
\centering
\resizebox{\textwidth}{!}{%
\begin{tabular}{lccccccccccc}
\toprule
\multicolumn{1}{c}{} &  & \multicolumn{5}{c}{{\color[HTML]{680100} \textbf{One-shot ICL}}} & \multicolumn{5}{c}{{\color[HTML]{303498} \textbf{Three-shots ICL}}} \\
\multicolumn{1}{c}{\multirow{-2}{*}{\textbf{Model}}} & \multirow{-2}{*}{\textbf{\#Parameters}} & {\color[HTML]{680100} \textbf{F1 Func.}} & {\color[HTML]{680100} \textbf{F1 Param.}} & {\color[HTML]{680100} \textbf{Part. Acc.}} & {\color[HTML]{680100} \textbf{Full Acc.}} & {\color[HTML]{680100} \textbf{Win Rate}} & {\color[HTML]{303498} \textbf{F1 Func.}} & {\color[HTML]{303498} \textbf{F1 Param.}} & {\color[HTML]{303498} \textbf{Part. Acc.}} & {\color[HTML]{303498} \textbf{Full Acc.}} & {\color[HTML]{303498} \textbf{Win Rate}} \\
\midrule
\textbf{xLAM-1b-fc-r} & 1B & 0.19 & 0.08 & 0.09 & 0.00 & 0.01 & 0.22 & 0.09 & 0.09 & 0.03 & 0.02 \\
\textbf{xLAM-2-1b-fc-r} & 1B & 0.41 & 0.13 & 0.13 & 0.00 & 0.00 & 0.43 & 0.12 & 0.13 & 0.00 & 0.00 \\
\textbf{xLAM-7b-fc-r} & 7B & 0.49 & 0.17 & 0.15 & 0.00 & 0.03 & 0.55 & 0.23 & 0.23 & 0.15 & 0.14 \\
\textbf{xLAM-2-8b-fc-r} & 8B & 0.48 & 0.15 & 0.14 & 0.00 & 0.01 & 0.47 & 0.17 & 0.15 & 0.04 & 0.04 \\
\textbf{Hammer2.0-7b} & 7B & 0.56 & 0.24 & 0.21 & 0.07 & 0.16 & 0.61 & 0.30 & 0.29 & 0.22 & \underline{0.25} \\
\textbf{Hammer2.1-7b} & 7B & 0.10 & 0.05 & 0.05 & 0.01 & 0.01 & 0.16 & 0.10 & 0.11 & 0.08 & 0.08 \\
\textbf{Llama-3-1-8B-Instruct} & 8B & 0.64 & 0.19 & 0.17 & 0.06 & 0.06 & 0.63 & 0.22 & 0.22 & 0.16 & 0.11 \\
\textbf{ToolACE-8B} & 8B & 0.43 & 0.13 & 0.13 & 0.00 & 0.00 & 0.50 & 0.15 & 0.13 & 0.00 & 0.00 \\
\textbf{ToolACE-2-Llama-3.1-8B} & 8B & 0.28 & 0.10 & 0.13 & 0.00 & 0.00 & 0.29 & 0.10 & 0.13 & 0.00 & 0.00 \\
\textbf{Granite-20B-FunctionCalling} & 20B & 0.64 & 0.20 & 0.17 & 0.02 & 0.05 & 0.61 & 0.25 & 0.26 & 0.21 & 0.20 \\
\textbf{Mixtral-8x7B-Instruct-v0.1} & 46.7B & 0.22 & 0.07 & 0.05 & 0.00 & 0.01 & 0.32 & 0.13 & 0.14 & 0.09 & 0.09 \\
\textbf{xLAM-8x7b-fc-r} & 46.7B & 0.40 & 0.15 & 0.16 & 0.01 & 0.01 & 0.43 & 0.16 & 0.17 & 0.02 & 0.03 \\
\textbf{Llama-3-1-70B-Instruct} & 70B & 0.41 & 0.19 & 0.15 & 0.04 & 0.09 & 0.33 & 0.17 & 0.15 & 0.07 & 0.11 \\
\textbf{Mixtral-8x22B-Instruct-v0.1} & 141B & 0.49 & 0.21 & 0.17 & 0.06 & 0.07 & 0.65 & 0.29 & 0.28 & 0.21 & 0.23 \\
\textbf{xLAM-8x22b-fc-r} & 141B & 0.53 & 0.21 & 0.22 & \underline{0.12} & 0.03 & 0.50 & 0.23 & 0.25 & 0.17 & 0.06 \\
\textbf{Llama-3-1-405B-Instruct-fp8 }& 405B & 0.41 & 0.14 & 0.08 & 0.03 & 0.10 & 0.41 & 0.18 & 0.13 & 0.07 & 0.14 \\
{\color[HTML]{242424} \textbf{DeepSeek-V3}} & {\color[HTML]{242424} 685B} &  \underline{0.69} &  \underline{0.36} &  \underline{0.27} & 0.09 & \underline{0.43} & \underline{0.69} & \textbf{0.42} & \underline{0.37} & \textbf{0.29} & \textbf{0.60} \\
\textbf{GPT-4o} (2024-08-06) & UNK & \textbf{0.73} & \underline{0.41} & \textbf{0.38} & \textbf{0.28} & \textbf{0.59} & \textbf{0.74} & 0.41 & \textbf{0.38} & \underline{0.28} & \textbf{0.60} \\
\bottomrule
\end{tabular}%
}
\caption{{ Evaluation results on \dataset{} on state-of-the-art LLMs with Direct Prompting technique. Models are sorted by their size. Experiments are done in one-shot and three-shot ICL settings. The best performance is highlighted in \textbf{bold}; the second best is \underline{underlined}. \textbf{Partial Sequence Accuracy (Part. Acc.)} denotes the percentage of calling the correct API sequence (API names and arguments), whereas \textbf{Full Sequence Accuracy (Full Acc.)} counts the percentage of times where the model gets the entire sequence of APIs correctly. Both scores range from 0 to 1. We also report \textbf{Win Rate}, which measures whether all the predicted APIs by the model are executable and lead to an exact match with the gold answer.}}
\label{tab:result}
\end{table*}

\subsection{Baselines}
 We extensively evaluated \dataset{} on 19 proprietary and open-source models, ranging in size from 1B to 685B parameters. This selection includes top tool-calling LLMs featured on the Berkeley Function-Calling Leaderboard (BFCL) \cite{berkeley-function-calling-leaderboard}, as well as state-of-the-art models known for strong function-calling capabilities. Among the tool-calling models, we include the xLAM \cite{zhang2024xlam, liu2024apigen}, Hammer \cite{lin2024hammer}, ToolAce \cite{liu2409toolace} model families, and Granite-20B-FunctionCalling \cite{abdelaziz2024granite}. We also evaluate a range of foundation models, including multiple sizes of LLAMA 3.1 \cite{dubey2024llama},  Mixtral\footnote{\url{https://huggingface.co/mistralai/}}, and DeepSeek-V3 \cite{guo2024deepseek}, and the state-of-the-art proprietary model GPT-4o \cite{hurst2024gpt}. To explore how an agentic LLM performs on \dataset{}, we also include AgentLM-13B \cite{zeng2023agenttuning}, which has been instruction-tuned using interaction trajectories from diverse agentic tasks.

\subsection{Experimental Settings}
The experiments are carried out with temperature 0.0 in one-shot and three-shot settings, i.e., the prompt contains one or three in-context learning (ICL) examples, respectively. To the best of our knowledge, all 18 open models were not trained with the \textit{label} assignment syntax in the output API sequence, so it was crucial to have ICL examples to get the best results. For each model, we used its specified prompt along with the special tags. Context length limitations prevented the inclusion of the entire API library in the prompt. Instead, we pre-processed the data to create a shorter API list for each sample. This list ensured the inclusion of the gold APIs, the APIs used in the in-context learning (ICL) examples, and some random APIs, keeping the total prompt length under 4K tokens. Output API calls were extracted from the model's response as a list of JSON objects, taking into account the specific prompt format and output structure for each model.
Finally, we evaluated a zero-shot ReAct \cite{yao2022react} agent with the best 4 open models based on the win-rate and AgentLM-13B, limiting max steps to 10.

\subsection{Metrics}

For a detailed evaluation, we use the following metrics: 1) F1 score for function and parameter names generation~\cite{abdelaziz2024granite}, 2) \textbf{\textit{Partial}} and \textbf{\textit{Full Match Accuracy}}, and 3) \textbf{\textit{Win Rate}}. 

LLM model response is a sequence of API calls, with each call consisting of an API name and its argument-value pairs. We use the \textbf{\textit{Partial Sequence Matching}} metric to determine how many predicted APIs (with their argument-value pairs) in a sequence match with the gold API sequence. In contrast, the \textbf{\textit{Full Sequence Matching}} metric evaluates whether the model predicts the exact full sequence of APIs, including both the API names and their argument-value pairs, when compared to the gold API sequence. In both cases, we calculate the scores for each sample and then compute the statistical mean across the entire dataset as the final score. We also use is the  \textbf{\textit{Win Rate}}, which measures if all the predicted APIs by the model are valid and when \textit{executed} lead to the gold answer. Unlike the \textit{F1} and \textit{Accuracy} measures, which assess the alignment of predictions with the gold API sequence, the \textit{Win Rate} focuses on the final gold answer. A win is recorded if the predicted answer matches the gold answer.

\subsection{Results} \label{sec:result}

Table~\ref{tab:result} presents a comparison of different baselines on the \dataset{} dataset with one-shot and three-shot ICL example settings. The low numbers of the best function calling models depict the complexity and toughness of the nested sequencing problem. GPT-4o and DeepSeek-V3 achieve the highest win-rate of 60\%, which is significantly below the acceptable numbers for real-world applications in general. This clearly depicts the significant scope for improvements for the models in various aspects of function calling, including nested sequencing. We inspected the models' outputs and identified several common issues across them. These models struggle with tasks such as assigning variables, utilizing output parameter details from the API specifications, and correctly passing variable names and corresponding output parameters to subsequent APIs, even with ICL examples\footnote{Note: While we acknowledge that these models were not trained using the robust data schema outlined in Section \ref{sec:data_schema}, the challenges associated with nested sequencing persist regardless of the schema used and remain an area where LLMs need improvement.}.

As anticipated, in most of our experiments in Table~\ref{tab:result}, the models are doing better across all the metrics when they are provided with three ICL examples in the prompt instead of one example. Across all models, \textit{Partial Sequence Match} scores are consistently higher than  \textit{Full Sequence Match} scores, which is expected, as the latter is a stricter metric than the former. In many cases, the \textit{Win Rate} is higher than the \textit{Full Accuracy} because models may follow alternative reasoning paths/trajectories to arrive at the correct final answer. While such deviations can penalize full or partial accuracy scores, they are still credited under the win rate for successfully reaching the gold answer. Hammer2.0-7b, despite being a smaller model, outperforms several larger tool-augmented LLMs. DeepSeek-V3 emerges as the strongest open-source model, closely matching GPT-4o's performance in the three-shot ICL setting, although it trails slightly in the one-shot setup. These results highlight that model size or architectural complexity is not the primary determinant of performance; rather, the ability to effectively follow instructions and leverage in-context examples plays a more critical role. This is evident as some large models like xLAM-8x22b-fc-r and Llama-3-1-405B-Instruct-fp8 underperform, while smaller models like Hammer2.0-7b achieve exceptional results.

\paragraph{ReAct-based Evaluation:} The previous results showed how poorly direct prompting of LLMs performed on \dataset{}. The literature has shown that agentic approaches have resulted in better performance on complex tasks. While many agentic architectures exist, we selected ReAct due to its popularity and its ability to reason over the output of the tool that is added to the prompt at each turn. 

\begin{table}[]
\centering
\resizebox{0.5\textwidth}{!}{%
\begin{tabular}{lcc}
\toprule
\textbf{Model} & {\color[HTML]{680100} \textbf{\begin{tabular}[c]{@{}c@{}}Direct Prompting \\ (One-shot ICL)\end{tabular}}} & {\color[HTML]{303498} \textbf{\begin{tabular}[c]{@{}c@{}}ReAct Agent \\ (Zero-shot)\end{tabular}}} \\
\midrule
\textbf{Hammer2.0-7b}  & \textbf{0.16} & 0.07 \\
\textbf{AgentLM-13B}  & 0.00 & 0.00 \\
\textbf{Mixtral-8x22B-Instruct-v0.1}  & 0.07 & \textbf{0.30} \\
\textbf{DeepSeek-V3}  & 0.43 & \textbf{0.46}\\
\bottomrule
\end{tabular}%
}
\caption{ Evaluation results (\textbf{Win Rate}) on \dataset{} comparing the performance of a ReAct Agent (zero-shot) to the Direct Prompting with a one-shot ICL example. For each model, the best performance is highlighted in bold for comparison.}
\label{tab:result_react}
\end{table}

Table~\ref{tab:result_react} summarizes the results of the ReAct agent~\cite{yaoreact} in compared to one-shot ICL direct prompting. Note that there was no ICL example provided in the ReAct case as the expected output does not need to follow the \textit{label} assignment syntax. We only report \textit{Win Rate}), which check if the trajectory of output tools lead to the gold answer, due to the single-step planning and execution approach of REACT as opposed to planning the entire API sequence at once in direct prompting. 

For larger models like Mixtral-8x22B-Instruct-v0.1 and DeepSeek-V3, the ReAct approach outperforms direct prompting, though there is still room for improvement. Notably, Mixtral-8x22B-Instruct-v0.1 shows the highest win-rate gain of 30\% with ReAct. Hammer2.0-7B performs better with direct prompting compared to the ReAct approach, Although AgentLM-13B is specifically trained for agentic tasks, it does not demonstrate strong performance on \dataset{}, indicating that agent-specific training or architectures do not always guarantee improved results in this setting. Output analysis reveals that the top-performing models exhibited better alignment with the ReAct format and occasionally relied on their parametric knowledge to replicate API functionalities. As a result, they achieved correct final outcomes, reflected in higher \textit{win-rate} despite inconsistencies in intermediate steps.

\subsection{Dataset Analysis}

To analyze the results, we model the samples in the \dataset{} as a Directed Acyclic Graph (DAG) where nodes are individual function calls and the edges are data dependencies between two nodes.

\begin{figure*}[t]
\centering
\begin{subfigure}{.33\textwidth}
  \centering
  \includegraphics[width=.95\linewidth]{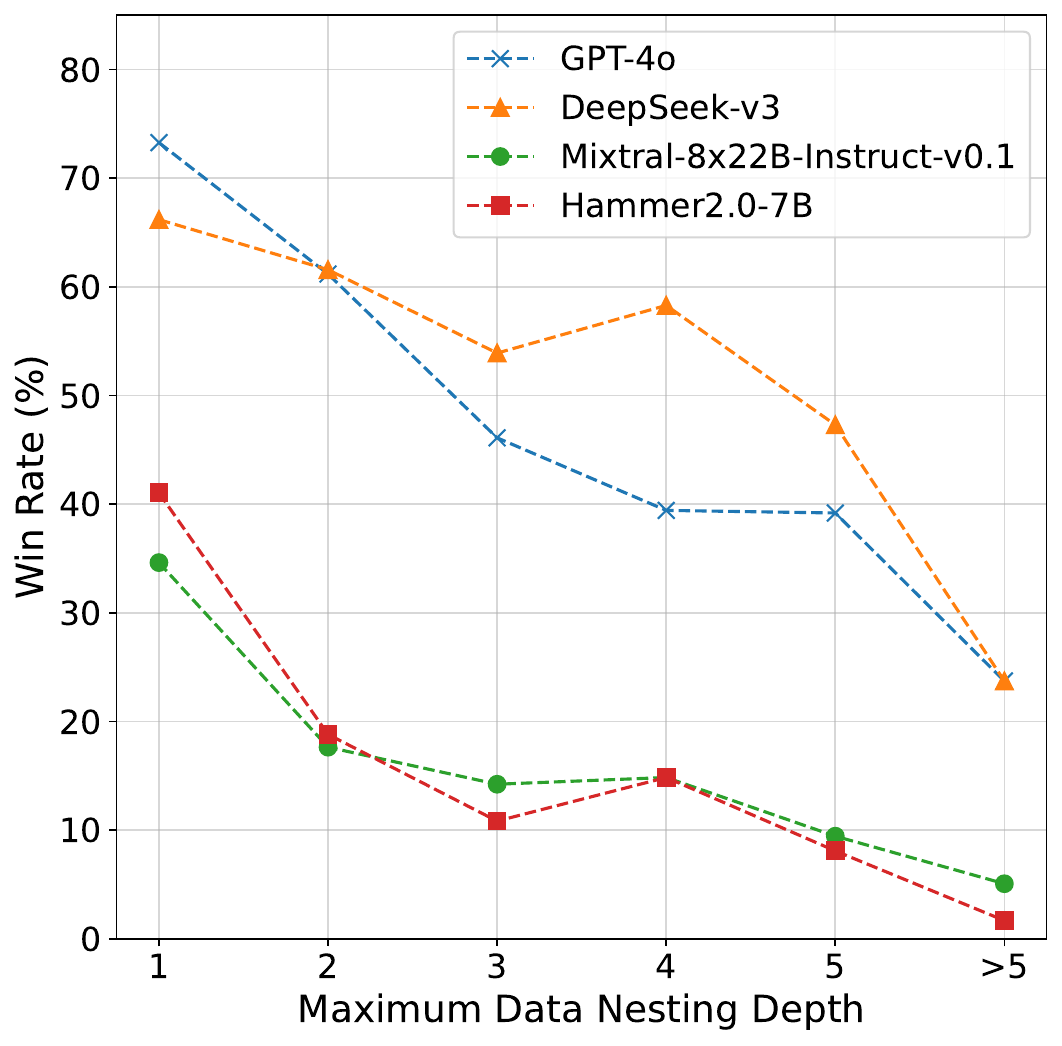}
  \caption{Data Nesting Depth Analysis}
  \label{fig:winrate_vs_nestingdepth}
\end{subfigure}%
\begin{subfigure}{.33\textwidth}
  \centering
  \includegraphics[width=.95\linewidth]{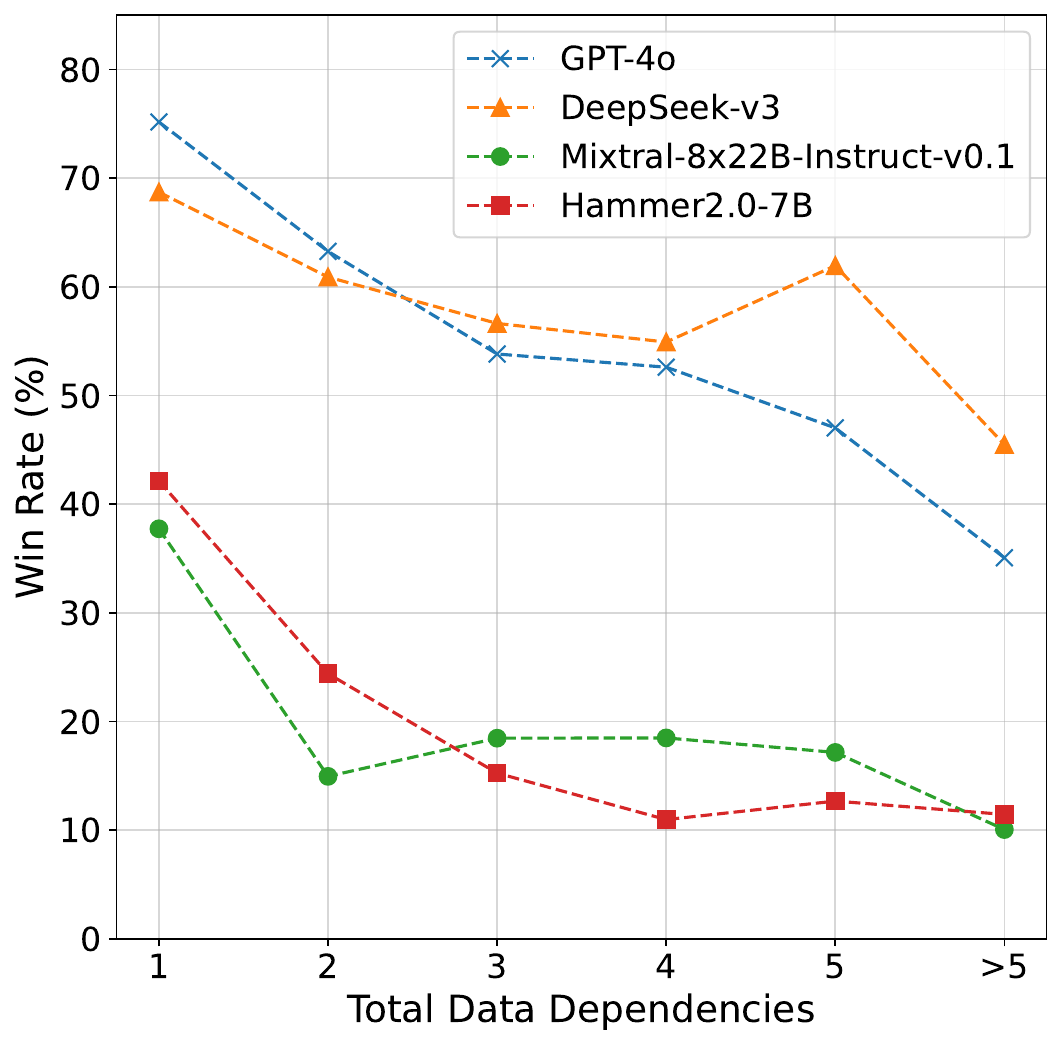}
  \caption{Total Data Dependencies Analysis}
  \label{fig:winrate_vs_ndependencies}
\end{subfigure}%
\begin{subfigure}{.33\textwidth}
  \centering
  \includegraphics[width=.95\linewidth]{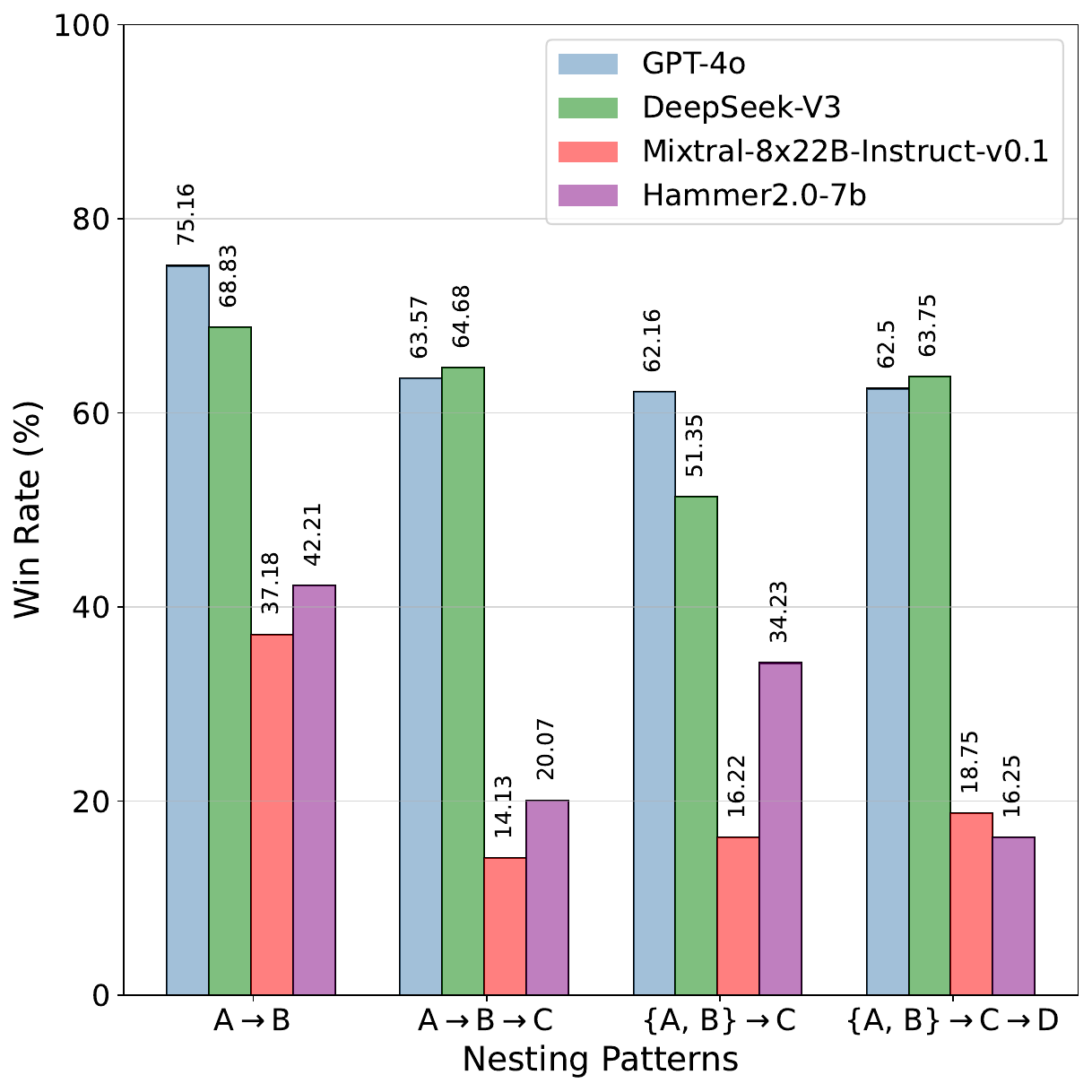}
  \caption{Nesting Pattern Analysis}
  \label{fig:winrate_vs_nestingpatterns}
\end{subfigure}
\caption{Top 4 models' performances (on the three-shot setting) with a varying number of (a) longest data dependencies, (b) the total number of data dependencies, and (c) common data nesting patterns in \dataset{}. We observe that models perform well for the simple data samples, however, the performance drops sharply with complicated structural patterns in the data.}
\label{fig:test}
\end{figure*}


\paragraph{Data Nesting Depth Analysis} In Figure \ref{fig:winrate_vs_nestingdepth}, we present the win rate for top-performing models against varying levels of maximum depth in the DAG structure, which corresponds to the longest nested data dependency flow in a sample. We observe that all models 
perform well for samples with maximum single nesting depth. However, the performance drops sharply with depths of two or more suggesting that long nested sequences present difficult scenarios for current models.

\paragraph{Total Data Dependency Analysis} In Figure \ref{fig:winrate_vs_ndependencies}, we present the win rate compared to the total number of data dependencies within a sample (a representation of the complexity of the sequence). The trends here are similar to the ones we observed in the Nesting Depth Analysis section. Models perform well for singular nested data dependency in the sample, achieving around 75\% \textit{win-rate}, but the performance drops sharply with two or more nested data dependencies.

\paragraph{Nesting Patterns Analysis} We identify common nesting patterns and analyze model performances for individual nesting patterns. Results are shown in Figure \ref{fig:winrate_vs_nestingpatterns}. We observe that for a simple pattern such as \texttt{A $\rightarrow$ B}, where the output of \texttt{A} is used as input by \texttt{B}, all models perform fairly well. However, for complex patterns such as \texttt{\{A, B\} $\rightarrow$ C}, where the outputs of both \texttt{A} and \texttt{B} are used as input by \texttt{C}, model performance decreases significantly. This suggests that models currently struggle with more complex patterns present in \dataset{}.


\section{Challenges}
\label{sec:challenges}
Results show that \dataset{} posed a challenge for state-of-the-art LLMs for several reasons. 

\smallskip
\noindent\textbf{Data-type Adherence for the Input/Output Parameters} ~  
In the API specification, we define the data type for all parameters. The `type' field specifies the data type, such as string, number, list, etc. Since APIs follow a strict structure for both input and output, it is crucial for the model to adhere to these specified formats. If the model fails to follow this, especially in cases involving nested functions where the output of one API is used as the input for another, the process will fail if the output type doesn’t match the expected input type. 


\smallskip
\noindent\textbf{Variable Assignments} ~
As discussed in Section \ref{sec:data_schema}, we add variable assignments for each API in the output to manage parallel function calls, which is very common in real-life applications. Below is an example of parallel nested function calls:

\begin{lstlisting}[language=json,firstnumber=1]
{"input": "What is the difference between the squares of 4 and 3?"
 "output": [
    {"name": "square", "arguments": {arg_0: 4}, 
      "label": "$var_1"}, 
    {"name": "square", "arguments": {arg_0: 3}, 
      "label": "$var_2"},
    {"name": "subtraction", "arguments": {arg_0: $var_1.result$, arg_1: $var_2.result$},  
      "label": "var_3"}, 
 ]}
\end{lstlisting}
The example highlights the complexity of distinguishing repeated functions with different outputs, which models struggle with due to a lack of schema training—a challenge also supported by our qualitative analysis in Section 4.4.


\smallskip
\noindent\textbf{Implicit API calling} ~
Implicit function calling occurs when a model must identify and invoke the appropriate APIs to solve a user query, even though the query doesn't explicitly mention them. This requires understanding the problem, selecting the correct functions, and filling in parameters using query details or previous outputs—adding significant complexity to the task. Figure \ref{fig:data_flow} demonstrates an example of implicit function calling, where the user query presents an arithmetic problem without explicitly stating the APIs involved.



\section{Conclusion}

In this work, we introduced \dataset{}, a new benchmark for evaluating the LLMs on nested sequences of API function calling. Existing LLMs perform poorly on this dataset than on existing benchmarks. We also studied their performance and identified several modes of failure. In addition, we outlined many challenges this dataset poses to LLM function calling approaches. By releasing this dataset under a permissive open-source license, we aim to advance tool calling and enable solutions to more realistic, challenging tasks.



\bibliography{custom}




\end{document}